\documentclass[letterpaper, 10 pt, conference]{IEEEconf}  

\IEEEoverridecommandlockouts                              

\usepackage{mathptmx} 
\usepackage{amsmath} 
\usepackage{amssymb}  
\usepackage{graphics} 
\usepackage{graphicx}
\usepackage[tight,footnotesize]{subfigure}
\long\def\invis#1{}
\usepackage{cite}
\usepackage{url}
\usepackage{hyphenat}
\usepackage{amsmath,systeme}
\usepackage{xcolor}
\usepackage[linesnumbered,ruled]{algorithm2e}
\usepackage{siunitx}
\usepackage{enumerate}
\usepackage{mathtools}
\usepackage{booktabs}
\usepackage{subcaption}

\usepackage[utf8]{inputenc}
\usepackage{ wasysym }
\usepackage{pgfplots}
\usepackage{stackengine} 
\usepackage{tikz}
\usepackage{gensymb}
\usepackage{pgfplotstable}
\usepgfplotslibrary{colormaps}
\pgfplotsset{colormap/hot}
\pgfplotsset{width=9cm,compat=1.9}
\pgfplotsset{layers/my layer set/.define layer set={background,main,foreground}{ },set layers=my layer set,}
\usepackage{xcolor}
\usepackage{tabularx}
\usepackage{multirow}
\usepackage{graphicx} 

\newcommand\bmat{\begin{bmatrix}}
\newcommand\emat{\end{bmatrix}}
\usepackage{textcomp}
\usepackage{svg}
\usepackage{algpseudocode}
\usepackage{hyperref}

\usepackage[compact]{titlesec}
\usepackage[font=small]{caption}

\def\floatsep{4.0pt}
\def\textfloatsep{0.0pt}
\def\dblfloatsep{4.0pt}
\def\dbltextfloatsep{4.0pt}

\usepackage{fp}
\FPset{\pb}{0}
\newcommand{\pagebudget}[1]{}

\usepackage{xcolor}

\newif\ifmarkchange

\definecolor{changed}{rgb}{0.0,0.0,0.0}
\ifmarkchange
\definecolor{changed}{rgb}{0.91, 0.05, 0.05}
\fi

\title{
\LARGE \bf SHRUMS: Sensor Hallucination for Real-time Underwater\\ Motion Planning with a Compact 3D Sonar}

\author{Author(s) ommited due to double-blind standards}
\author{Susheel Vadakkekuruppath$^1$, Lauritz R. Fosso$^2$, Michail Chatzispyrou$^3$, Herman B. Amundsen$^4$, \\Sveinung J. Ohrem$^2$, Jason M. O'Kane$^1$, and Marios Xanthidis$^2$
\thanks{$^1$ Department of Computer Science and Engineering, Texas A\&M University, USA, {\tt\small [susheelvk, jokane]@tamu.edu}}
\thanks{$^2$ Maritime Autonomy and Robotics, SINTEF Ocean, {\tt\small [lauritz.fosso, sveinung.ohrem, marios.xanthidis]@sintef.no}}
\thanks{$^3$ Michail Chatzispyrou is with the Department of Mechanical Engineering, University of Delaware, USA, {\tt\small michalis@udel.edu}}
\thanks{This work was supported by the Research Council of Norway (EchoNav: NO-359447, NG-UI: NO-364227) and National Science Foundation (2313928)}
}

\begin{document}

\maketitle
\thispagestyle{empty}
\pagestyle{empty}

\begin{abstract}
Autonomous navigation in 3D is a fundamental problem for autonomy.
Despite major advancements in terrestrial and aerial settings due to improved range sensors including LiDAR, compact sensors with similar capabilities for underwater robots have only recently become available, in the form of 3D sonars.
This paper introduces a novel underwater 3D navigation pipeline, called SHRUMS (\underline{S}ensor \underline{H}allucination for \underline{R}obust \underline{U}nderwater \underline{M}otion planning with 3D \underline{S}onar).
%
To the best of the authors' knowledge, SHRUMS is the first underwater autonomous navigation stack to integrate a 3D sonar and demonstrate safe operation in complex 3D environments without reliance on visual input, thereby remaining independent of visibility conditions.
To accommodate the intricacies of the novel sensor data stream while achieving real-time locally optimal performance, SHRUMS introduces the concept of hallucinating sensor measurements from non-existent sensors with convenient arbitrary parameters, tailored to application specific requirements.
%
%
The proposed concepts are validated in simulations informed with real 3D sonar sensor data, in-pool clean-water trials, and field deployments in real conditions with significantly degraded visibility conditions.
%

\end{abstract}



\section{Introduction}
\label{sec:intro}

\begin{figure}[ht]
    \centering
    \begin{tabular}{c}   
        \subfigure[]{\includegraphics[width=0.45\textwidth]{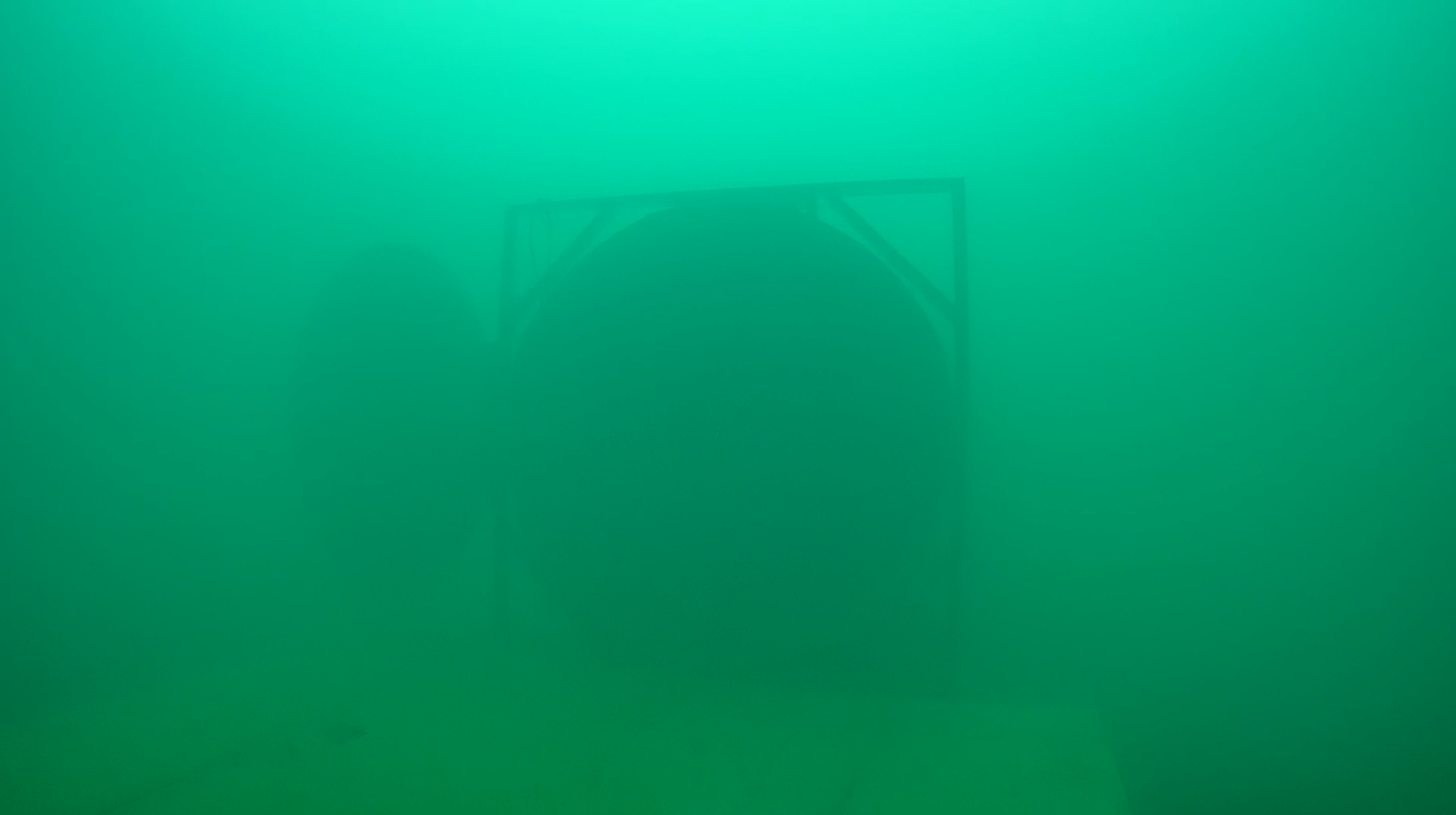}} \\[0.5em] 
        \subfigure[]{\includegraphics[width=0.45\textwidth]{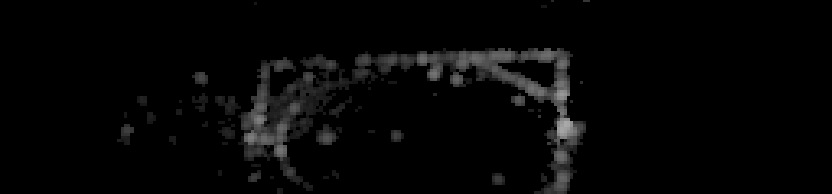}} \\[0.5em]
    \end{tabular}
    \begin{tabular}{cc}   
        \subfigure[]{\includegraphics[width=0.2\textwidth, trim=4cm 0cm 7cm 1cm, clip=true]{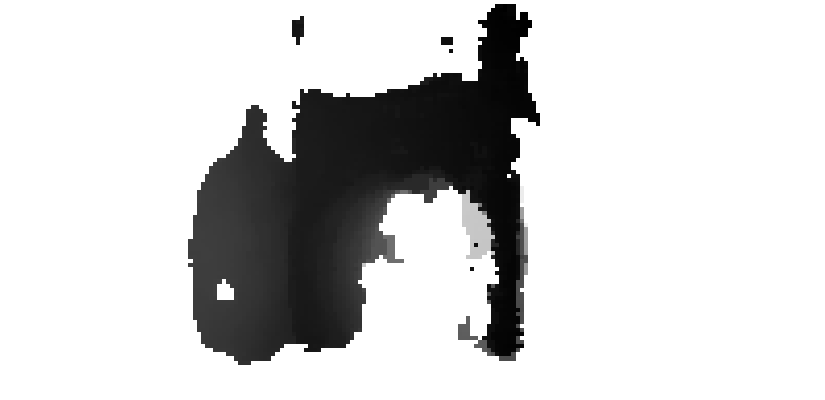}} &
        \subfigure[]{\includegraphics[width=0.225\textwidth]{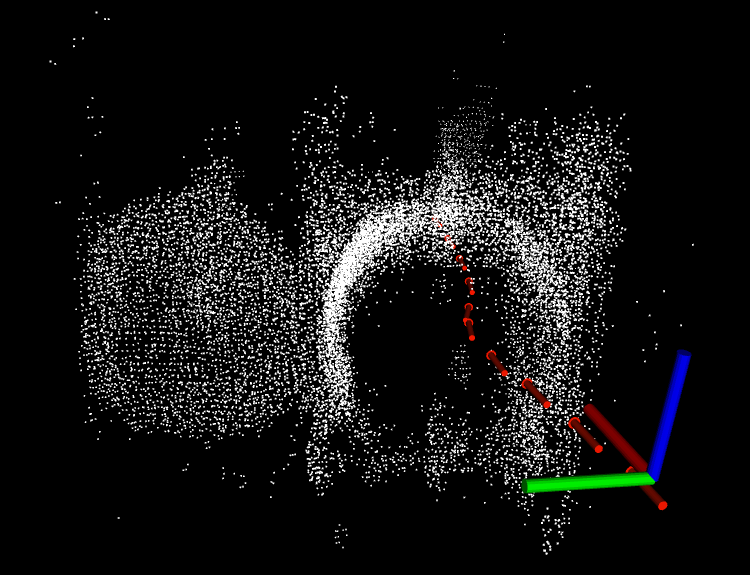}}
    \end{tabular}
    \caption{(a) A submerged container as seen in a video from an underwater robot's camera.  These extremely poor visibility conditions severely limit vision-based methodologies. (b)~A range image from the Sonar 3D-15, showing significantly clearer information about the container's geometry.  (c)~An example range image produced using sensor hallucination within SHRUMS (the proposed pipeline), enhancing the characteristics of the original sensing capacity in terms of resolution, field of view, and sensing completeness. (d)~A 3D path produced by SHRUMS to navigate the observed map.}
    \label{fig:beauty}
\end{figure}

Underwater autonomy is a field of rapidly growing importance, due to both strong industrial interest in resource utilization, public safety, and infrastructure inspection and maintenance~\cite{yoerger2007techniques, nauert2023inspection}, and academic interest, due to its unique challenges in sensing and acting~\cite{JoshiIROS2019, tugal2022sliding} and for its multidisciplinary applications such as marine archaeology~\cite{diamanti2024advancing}.
Interestingly, fundamental problems such as simulateous localization and mapping (SLAM) and autonomous navigation, which are often considered `solved' in ground~\cite{yang20223d}, aerial~\cite{papachristos2019autonomous}, and space robotics~\cite{amzajerdian2023overview}, are open problems in the underwater domain.  
These persistent challenges are, in part, due to poor propagation of electromagnetic waves, on which the state-of-the-art solutions rely for obtaining range measurements.
A common denominator is the utilization of 3D range sensors, such as LiDARs, as the basis for many approaches to state estimation and spatial understanding.

Though acoustic sensors such as multibeam and side-scan sonars generally perform well underwater, they provide, at best, information about only a 2D slice of the environment.  
This paucity of information has historically limited autonomous navigation to fly-overs in obstacle-free water columns or at best 2.5D motions~\cite{fallon2011efficient,hernandez2019online}.
The limited prior work on underwater autonomous 3D navigation generally operates in simulated and known environments~\cite{hernandez2019online,hover2012advanced,xanthidis2020navigation}, or is strongly limited in applicability by a reliance on visual sensing~\cite{xanthidis2020navigation,manderson2020vision}.
Unfortunately, vision-based sensing can be rendered completely ineffective in zero-visibility conditions --- a common occurrence in benthic environments and during inspections of infrastructure, as shown in Figure~\ref{fig:beauty}-a. 

This bottleneck has been identified by emulating the volumetric information from range sensors, such as LiDARs, relying on imaging sonar measurements~\cite{jaber2025mv3d}, which result in challenging-to-resolve spatial ambiguities, and configurations that combine camera and sonars~\cite{yang2022monocular} or multiple multi-beam sonars~\cite{mcconnell2020fusing}, which result in limited FOVs. 
To address such issues, mechanical scanning 3D sonars have been presented in the past, though they require static environments and limited motion during operation. These constraints prohibit utilization on mobile AUVs performing autonomous navigation.
Previous studies on point cloud registration with older 3D sonars, capturing the 3D structure directly by producing 3D point clouds, have been presented~\cite{ferreira20223dupic}, though the tested hardware was unsuitable for most mobile underwater platforms due to the cost, energy, and size.
However, the recent introduction of compact 3D sonars, such as the Sonar 3D-15 from Water Linked\footnote{\url{https://waterlinked.com/product/sonar-3d-15/}}, provides new opportunities to enable underwater autonomy, as recent research has shown~\cite{burgul2025underwater,archieri2026insso3d}.  See Figure~\ref{fig:beauty}b.

This paper presents an autonomous real-time 3D navigation pipeline  called SHRUMS (\underline{S}ensor \underline{H}allucination for \underline{R}obust \underline{U}nderwater \underline{M}otion planning with 3D \underline{S}onar), which utilizes a compact 3D sonar.
In contrast to previous methodologies, SHRUMS provides safe 3D paths utilizing exclusively acoustic 3D sensing with no reliance on optical feedback, providing robustness against challenging visibility conditions, and allowing safe operations even in zero-visibility. 
SHRUMS is designed to be resilient against unexpected tracking losses due to localization or map drifts, by exploiting spatial local consistency and having the capacity to quickly rebuild a map representation 
and plan a 3D trajectory in real-time.
In addition, the produced trajectories are locally optimal, continuous-time safe, and consider the geometry of the surroundings captured with no explicit dependency on a volumetric map resolution.
To the best of our knowledge, this is the first pipeline to utilize 3D sonar for robust autonomous 3D navigation with such strong properties.
A related thread in the aerial robotics community has proposed real-time methodologies for autonomous navigation.
FASTER~\cite{tordesillas2021faster} incrementally refines trajectories within convex corridors to guarantee safety, while EGO-Planner~\cite{zhou2020ego} leverages gradient-based optimization on voxelized occupancy maps. 
Although both achieve real-time replanning once initialized, their reliance on start-up computations can hinder immediate responsiveness in previously unmapped regions. 
Moreover, the dependence on voxel-based representations introduces resolution dependencies and discretization artifacts that limit geometric fidelity in cluttered environments.
This not only limits the accuracy of the geometric representation of the environment, but also generally requires a high-resolution sensor that 
will adequately cover the surrounding voxels.

The SHRUMS pipeline is able to efficiently mitigate intricacies of 3D sonar data using a form of \textit{sensor hallucination}, inspired by works that have hinted at the utility of altering sensing measurements to hallucinate obstacles for safety negotiations in multi-robot systems~\cite{xiao2021toward,park2022learning}, inflate obstacles in the latent space for safe avoidance~\cite{kulkarni2023task}, and offer alternative informative out-of-body perspectives~\cite{abdullah2024ego}.
The key idea is to avoid computationally expensive processing of the entire local map (as required in previous methodologies with mesh reconstruction in unstructured point clouds~\cite{xanthidis2020navigation}). 
 Instead, SHRUMS enhances sensing capacity by simulating inputs of non-existing sensors of arbitrary properties, placement, and configuration, using the local map to inform the response of the real system. 
 Figure~\ref{fig:beauty}c showcases the input improvement with sensor hallucination and Figure~\ref{fig:beauty}d a path instance navigating this scene in 3D.
This approach enables readiness to handle tracking losses, accelerates the map computation of the surroundings, and compensates for potential sensor noise or incompleteness. 

In short, the main contributions of this work are:
\begin{itemize}
    \item SHRUMS, a novel real-time underwater 3D navigation pipeline which can operate in unknown environments and arbitrary visibility conditions, utilizing a compact 3D sonar for the first time.
    \item \textit{Sensor hallucination}, a novel concept that bridges the gap between direct methods that react to single measurements and indirect approaches that construct environment representations over time.
    This improves computational efficiency, sensing information, and methodology adaption in semi-direct methods, formulated in a comprehensive way.
    \item Extensive validation through realistic simulations, in-pool testing, and field deployments.
\end{itemize}

\section{Problem Statement}
\label{sec:prob}
The goal of SHRUMS is to provide locally-optimal autonomous 3D navigation with obstacles for autonomous underwater vehicles (AUVs), using a 3D range sensor.
Consider an AUV with state $s = (x,y,z,\theta,\phi,\psi)$, in which $x$, $y$, and $z$ correspond to the translational 3D coordinates and $\theta$, $\phi$, $\psi$ correspond to the orientation given by roll, pitch, and yaw, respectively.
Let $s_1$ denote the initial pose of the AUV and let $G$ denote a closed goal region around the goal pose $s_g \in G$.
The robot must avoid a set of $m$ static obstacles $O=\{o_1, o_2, \dots, o_m\}$, with $A_s$ be the volume occupied by the robot at state $s$ and $A_o$ the volume occupied by obstacle $o$.
To plan its motions, the robot utilizes a sequence of consecutive range measurements $C^{T'}= \{c_{t_i}, c_{t_{i+1}}, \dots, c_{t_{i+m}}\}$.
Each range measurement in $C^{T'}$ is a point cloud observing a subspace of the surfaces of the obstacle set $O$. 

Then, given the measurement history $C^{T}$, a solution is a path $S = \{ s_1, \ldots, s_n \}$ to the optimization problem:
    \begin{align}
        f(S|C^{T}) &= \displaystyle\min_{s_i \in S}
        \sum\limits_{i=1}^{n-1} \left \| s_{i+1}-s_i \right \|, 
        \label{eq:obj1} \\
        \textrm{s.t.}~& 
        \left(\bigcup\limits_{s\in S} A_{s} \right)
        \cap
        \left( \bigcup_{o \in O} A_o \right)
        = \varnothing
        \label{eq:obj2}
    \end{align}
Simply put, given range measurements $C^{T}$, the solution is a minimum-displacement collision-free path $S$ to the goal $G$.

\section{Method Description}
\label{sec:app}
\begin{figure*}[t]
    \hspace*{-1.1cm}
    \includegraphics[width=2.25\columnwidth, trim=0cm 2cm 0cm 2cm, clip=true]{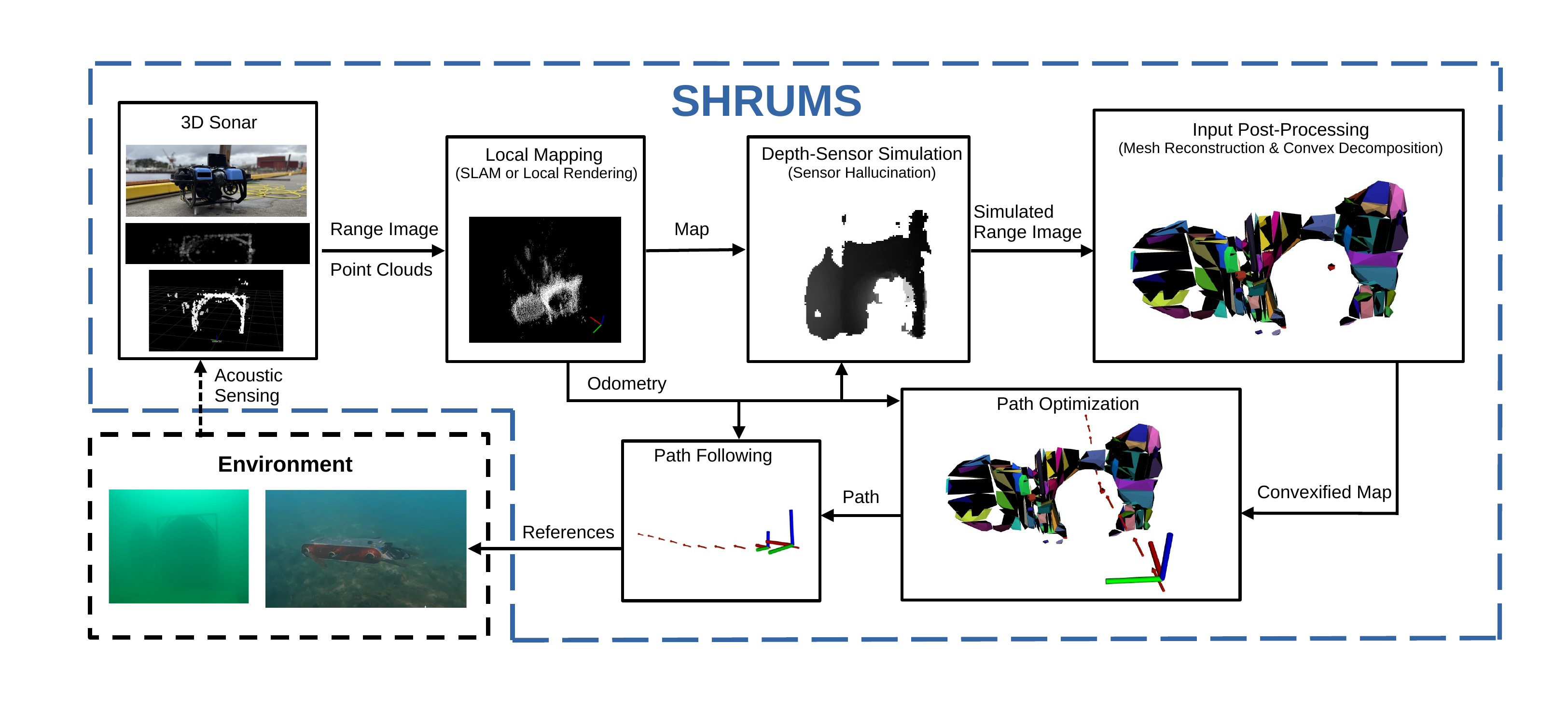}
    \caption{The SHRUMS pipeline with its main processing stages, including the integration of a 3D Sonar, local mapping, sensor hallucination, mesh reconstruction and convex decomposition, path optimization, and, finally, the path follower informing the robot's controls.}
    \label{fig:pipeline}
\end{figure*}

\subsection{Preliminaries on Water Linked's Sonar 3D-15}
\label{sec:sonar}
For the development of the proposed pipeline and our testing we utilized the newly introduced Sonar 3D-15 by Water Linked. 
The sensor has a maximum range of 15\si{m},  a variable field-of-view (FOV) of $90^\circ(\pm 45^\circ)\times45^\circ(\pm 29^\circ)$, an angular resolution of $0.35^\circ\times0.60^\circ$, and an image size of $256\times64$ in the horizontal and vertical directions respectively.
Along with the range image, the sensor also outputs a point cloud of detected structures at around 5-6 Hz. 
The sensor's acoustics operate at a high frequency of at least $1.2$\si{MHz}.
%

\begin{figure}[t]
    \centering
    \begin{tabular}{cc}   
        \subfigure[]{\includegraphics[width=0.45\columnwidth]{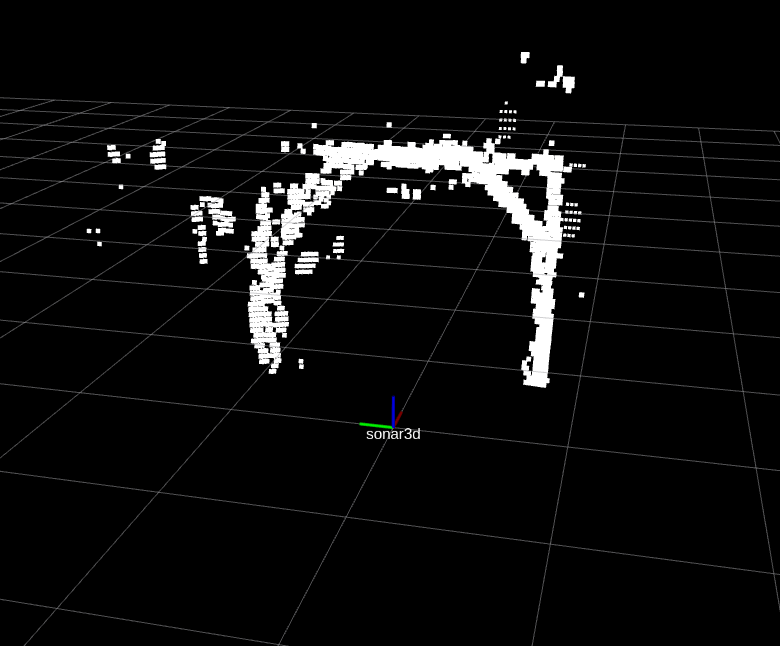}} &
        \subfigure[]{\includegraphics[width=0.45\columnwidth]{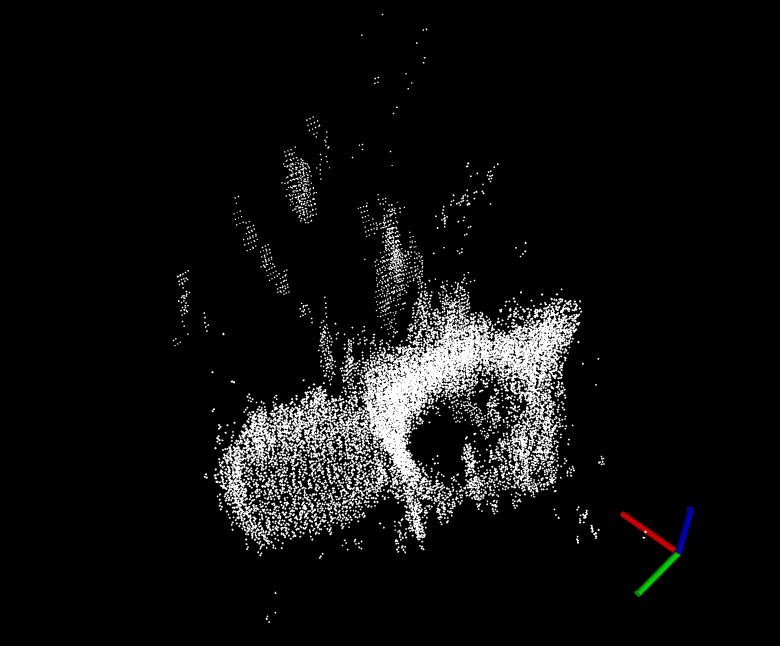}} 
    \end{tabular}
    \caption{Intricacies of the 3D sonar's sensing output. (a) A point cloud from the 3D sonar, though there is incomplete coverage of some surfaces, such as the door at the left side. (b) Local consistency is exploited by KISS-ICP to provide the entire scene.}
    \label{fig:sonar_issues}
\end{figure}

A core observation for our experience is that a single frame often did not contain dense information for the entirety of the observable scene, as with commonly used in-air range sensors.
Despite this limitation, the measurements were consistent enough and aggregating a few consecutive frames provided good coverage of the surrounding at satisfactory density for capturing geometric details of the entire scene.
This key attribute drove the design of the proposed pipeline towards a semi-direct method, given the necessity of quickly producing and maintaining local maps that provide better coverage of the observable scene, shown in Figure~\ref{fig:sonar_issues}.

\subsection{Sensor Hallucination and Proposed Pipeline}

The goal of SHRUMS is to provide an underwater autonomous navigation pipeline that allows for obstacle avoidance in 3D, avoids explicit geometric simplifications of the surroundings and the robotic platform, provides continuous time safety to ensure safe paths at arbitrary resolution, and computes paths in real-time.
For this purpose, inspired by previous autonomous underwater navigation frameworks~\cite{xanthidis2020navigation}, SHRUMS borrows the path-optimization formulation of Trajopt~\cite{schulman2014motion}, which, instead of using state-to-obstacles collision constraints, utilizes the swept-out-volume of the transitions by forming a convex set between each pair of consecutive states.
This formulation necessitates that the obstacle space is described as a set of convex polytopes.

As indicated in previous pipelines~\cite{xanthidis2020navigation}, point clouds can be processed on-line to produce convex polytopes, by creating a mesh and then performing convex decomposition. 
Unfortunately, both operations, particularly the mesh reconstruction on unstructured point clouds, are computationally demanding, making real-time replanning challenging due to the excessive computations needed to retrieve the spatial relationships between points.
In contrast, if a depth image were provided, that would correspond to an organized point cloud, and meshes could be created far more efficiently by limiting the distance queries in the local pixel neighborhood.

Given the aforementioned information and the intricacies of the sensor, the conflicts between safety and computational efficiency generate a new conundrum: In order to improve the mesh reconstruction computation, a depth image needs to be used, while simultaneously using the raw (or even processed) depth image directly from the sonar may compromise safety due to instances of not detecting all surfaces. 

We resolve this conundrum by introducing the concept of \textit{sensor hallucination}, by simulating the range data from a non-existing convenient sensor with desired user-defined placement, properties and configuration.
Such simulated range sensor inputs could have variable: 
\begin{itemize}
    \item placement, providing better perspectives for obstacle avoidance than the actual sensors, potentially even with more informative out-of-body third-person perspectives. 
    \item field of view (FOV), improving the perspective of sensors with narrower FOVs or focusing on more relevant areas for navigation from sensors with wider FOVs.
    \item resolution, adjusting sensing quality to computational needs or to match desired target values --- this aspect could enable easy adaption with transfer learning by emulating the sensing configuration of the training set.
    \item sensing distance, for emphasizing features closer in the local area or for providing more information with a larger range if a prior map is available.
    \item fundamental properties, such as improving the sensing quality of the original sensor, by utilizing a local map produced over multiple frames to inform planning instead of the direct input from other sensors with sensing intricacies such as 3D sonars.
    \item options not expanded in this work, such as utilizing a local map produced by multiple heterogeneous sensors, performing sensor fusion projected on a single input, or filtering out or adding obstacles in the scene for targeted applications.
\end{itemize}

Thus, by applying the concept of sensor hallucination in an autonomous underwater navigation framework using a 3D sonar, the proposed pipeline is shown in Figure~\ref{fig:pipeline}. 
In summary, SHRUMS consists of four consecutive steps that are parallelized as a pipeline:
\begin{enumerate}
    \item \textit{State Estimation}: The 3D sonar's range measurements inform a local state-estimation method to provide a local map of the surroundings and on-demand localization.
    \item \textit{Depth-Sensor Simulation}: Sensor hallucination is applied using the constructed local map to provide range data from a user-defined simulated sensor.
    \item \textit{Input Postprocessing}: A map representation is constructed for the motion planner, relying on fast mesh reconstruction and convex decomposition.
    \item \textit{Path Optimization}: Optimization-based motion planning is employed to produce real-time 3D solutions. 
    \item \textit{Path Following}: A simple path follower is applied to drive the robot along the produced path.
\end{enumerate}

The following subsections provide more details on the individual stages of this pipeline. 

\subsection{Local Mapping}
To overcome the sensor's inability to provide a full reconstruction of obstacles in a single scan, a SLAM methodology can be employed~\cite{burgul2025underwater,archieri2026insso3d}.  This SLAM method must be readily applicable to the sensor's output, able to efficiently and correctly register features between point clouds, and operate based on only limited numbers of measurements.
%
%
The need for SLAM that works with only few measurements is particularly crucial for enabling any-time obstacle avoidance, during or after tracking loss from the main SLAM module, with similar motivation as other works~\cite{nguyen2022motion}.
A family of SLAM methodologies following all these specifications exploits iterative closest point (ICP)-based scan matching, given its efficiency, minimalistic nature requiring only point clouds, its general capacity for sufficient data registration, and the fact that an informative local map can be constructed from as little as a single measurement.  
On the other hand, assuming locally consistent localization provided by other sensors (e.g. Doppler velocity logs (DVLs) and inertial measurement units (IMUs)), local scan registration for our purposes can perform also well, albeit with more noise and outliers.
We show that with SHRUMS, both options are viable, by utilizing KISS-ICP~\cite{vizzo2023ral} as a simple standard method for the first option, and local scan registration for the latter.
Figure~\ref{fig:sonar_issues}b shows an example of the former.

\subsection{Depth-Sensor Simulation}
To enable real-time sensor hallucination, the local map in the form of an unstructured point cloud was used to construct a depth image of desired specifications and placement.
Our implementation scales linearly to the number of points in the map, though more efficient implementations can be applied. 

If \rotatebox{30}{c} is the simulated hallucinated sensor, given a width $W_{pix}^{\rotatebox{30}{$c$}}$ and height $H_{pix}^{\rotatebox{30}{$c$}}$ in pixels for the desired image size, the horizontal $W_{FOV}^{\rotatebox{30}{$c$}}$ and vertical $H_{FOV}^{\rotatebox{30}{$c$}}$ angles defining its FOV, along with the minimum $d_{min}^{\rotatebox{30}{$c$}}$ and maximum $d_{max}^{\rotatebox{30}{$c$}}$ range for the sensor, a 3D point cloud $P=\{p_1, p_2, \dots, p_l\}$ of $l$ 3D points, and a desired pose $s$, then the intrinsics of the simulated sensor can be computed:
\begin{equation}
\begin{aligned}
f_x^{\rotatebox{30}{$c$}} &= \frac{W_{pix}^{\rotatebox{30}{$c$}}}{2\tan\!\left(\tfrac{W_{FOV}^{\rotatebox{30}{$c$}}}{2}\right)}, 
&\quad f_y^{\rotatebox{30}{$c$}} &= \frac{H_{pix}^{\rotatebox{30}{$c$}}}{2\tan\!\left(\tfrac{H_{FOV}^{\rotatebox{30}{$c$}}}{2}\right)}, \\
c_x^{\rotatebox{30}{$c$}} &= \tfrac{W_{pix}^{\rotatebox{30}{$c$}}}{2}, 
&\quad c_y^{\rotatebox{30}{$c$}} &= \tfrac{H_{pix}^{\rotatebox{30}{$c$}}}{2}.
\end{aligned}
\end{equation}
Then, the calibration matrix becomes:
\begin{equation}
K^{\rotatebox{30}{$c$}} =
\begin{bmatrix}
f_x^{\rotatebox{30}{$c$}} & 0 & c_x^{\rotatebox{30}{$c$}} \\
0 & f_y^{\rotatebox{30}{$c$}} & c_y^{\rotatebox{30}{$c$}} \\
0 & 0 & 1
\end{bmatrix}.
\end{equation}
If $T(s)$ is the transformation corresponding to pose $s$, then every point $p_i$ could be transformed into the sensor's reference frame:
\begin{equation}
\tilde{p}_i^{\rotatebox{30}{$c$}} = (T(s)^{-1}p_i)_{1:3}, \quad
\tilde{p}_i^{\rotatebox{30}{$c$}} =
\begin{bmatrix}
x_i^{\rotatebox{30}{$c$}} \\
y_i^{\rotatebox{30}{$c$}} \\
z_i^{\rotatebox{30}{$c$}}
\end{bmatrix}, \quad x_i^{\rotatebox{30}{$c$}} > 0.
\label{eq:coord}
\end{equation}
For simplification, we assume a pinhole camera model, which has a FOV less than $180^\circ$, thus, for efficiency, points with negative $x$ coordinate are discarded. 
Note that the overall concept can be applied with different sensor models. 
Moreover, larger FOVs could be achieved with more than one simulated sensor with tangent but non-overlapping views providing the desired combined coverage.

\begin{figure}[t]
    \centering
    \begin{tabular}{cc}
        \multicolumn{2}{c}{\subfigure[]{\includegraphics[width=0.95\columnwidth]{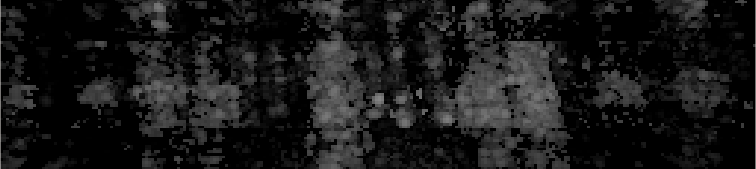}}} \\
        
        \subfigure[]{\includegraphics[width=0.45\columnwidth, trim=7.6cm 0cm 0cm 0cm, clip=true]{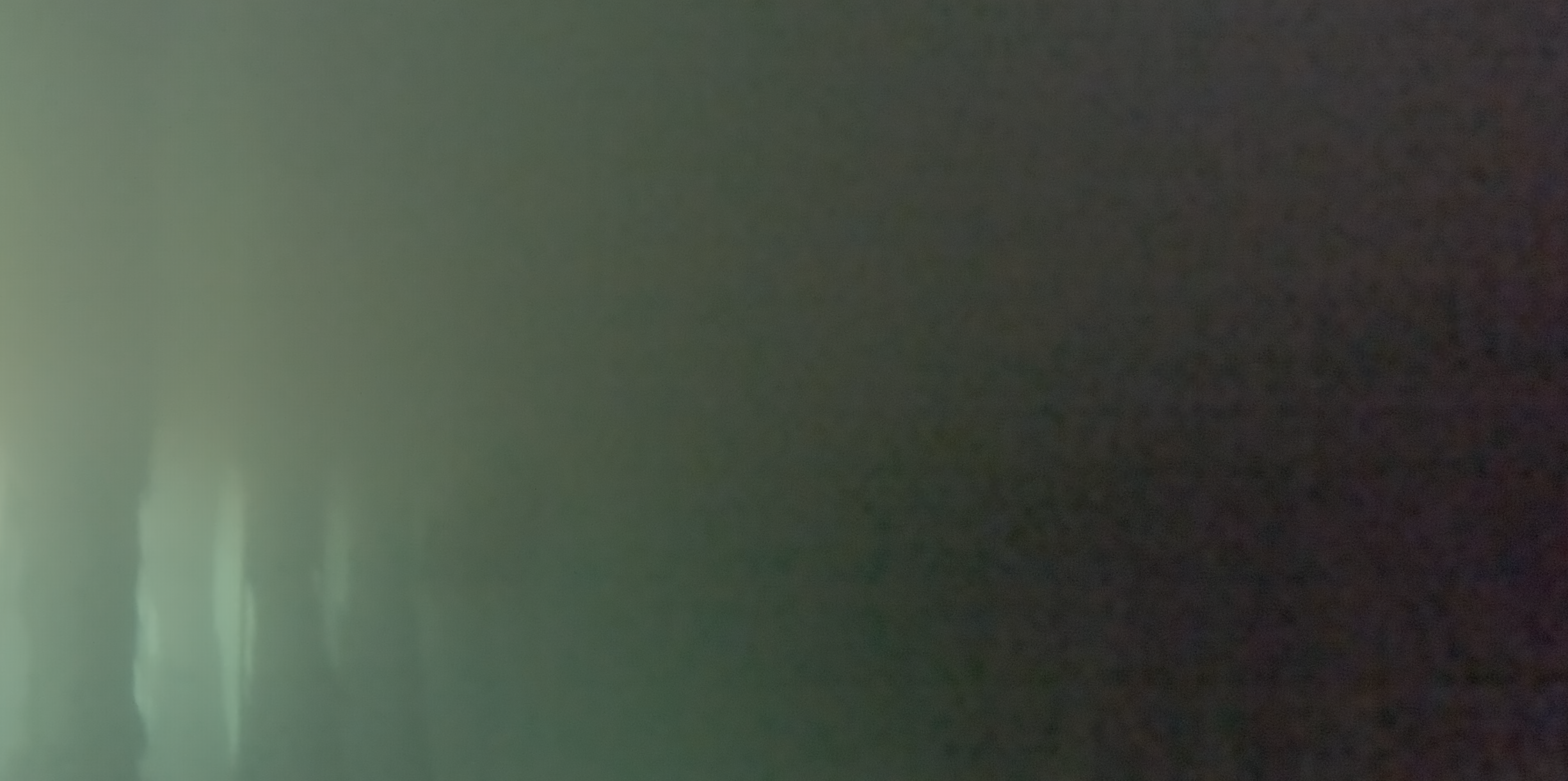}} &
        \subfigure[]{\includegraphics[width=0.45\columnwidth]{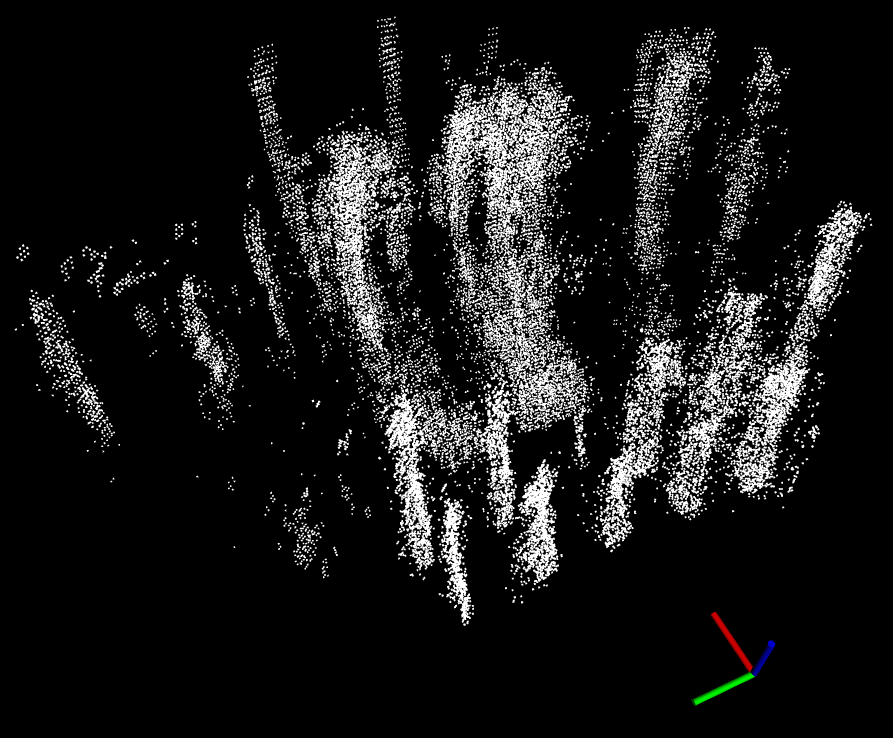}} \\
        
        \subfigure[]{\includegraphics[width=0.45\columnwidth]{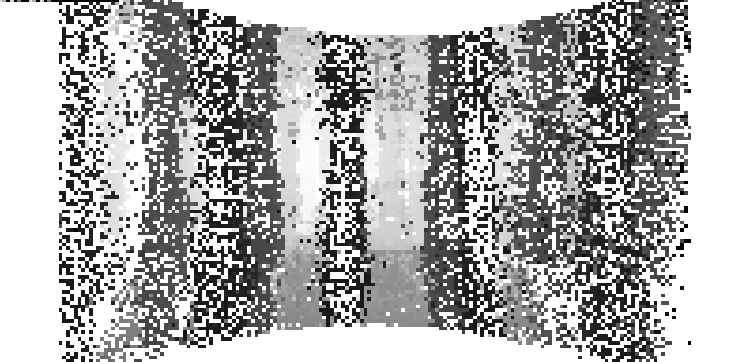}} &
        \subfigure[]{\includegraphics[width=0.45\columnwidth]{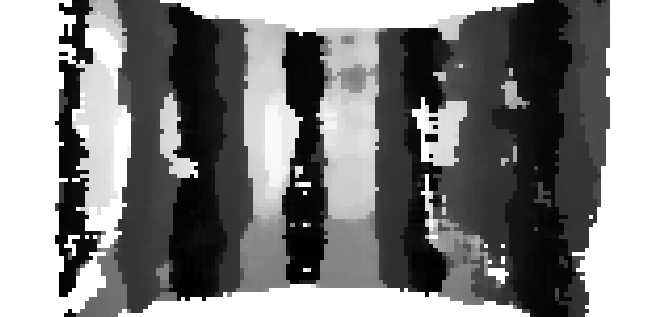}}
    \end{tabular}
    \caption{In (a), the sonar output is shown observing pillars from a deployment under a dock, while (b) shows the useless view from the robot's camera and (c) the local map produced by KISS-ICP. Simulated range images are shown in (d) before and (e) after applying a median filter. For higher-resolution images, dense range images may require preprocessing for smooth mesh reconstruction.}
    \label{fig:median}
\end{figure}

To continue, if $u$ is the horizontal and $v$ the vertical image coordinates, every point given (\ref{eq:coord}) is projected:
\begin{equation}
u_i^{\rotatebox{30}{$c$}} = f_x^{\rotatebox{30}{$c$}} \frac{-\,y_i^{\rotatebox{30}{$c$}}}{x_i^{\rotatebox{30}{$c$}}} + c_x^{\rotatebox{30}{$c$}}, 
\quad
v_i^{\rotatebox{30}{$c$}} = f_y^{\rotatebox{30}{$c$}} \frac{-\,z_i^{\rotatebox{30}{$c$}}}{x_i^{\rotatebox{30}{$c$}}} + c_y^{\rotatebox{30}{$c$}}.
\end{equation}
Finally, the value of every pixel $I(u,v)$ is computed given a corresponding set of points $\tilde{P}^{\rotatebox{30}{$c$}}(u,v)$:
\begin{equation}
I(u,v) =
\begin{cases}
\min\limits_{\tilde{p}_i^{\rotatebox{30}{$c$}} \in \tilde{P}^{\rotatebox{30}{$c$}}(u,v)} 
x_i^{\rotatebox{30}{$c$}}, 
& d_{min}^{\rotatebox{30}{$c$}} \leq x_i^{\rotatebox{30}{$c$}} \leq d_{max}^{\rotatebox{30}{$c$}}, \\[6pt]
\text{undefined} & \text{otherwise}.
\end{cases}
\end{equation}
Notice that when the proposed sensor is close to the local map's point cloud or at high resolutions, the depth image may be less dense than desired, leaving some cells with undefined values, given the mismatch with the point cloud's effective resolution.
As shown in Figure~\ref{fig:median}, without excluding alternative solutions, the simulated input was improved with the application of a median filter.

\subsection{Input Post-Processing}
In this step, the input from the simulated sensor is utilized to construct a map that can be used directly by the motion planner. 
To produce geometric simplifications that preserve the overall geometric complexity of the surroundings while avoiding volumetric partition, this work focuses on producing convex polytopes that can be readily used by path optimization for quick computation of distance queries between 3D volumes. 
This process involves two generally computationally expensive operations, in the context of real-time operation: Mesh reconstruction and convex decomposition.

Mesh reconstruction, in our experience and testing, was the most computationally expensive operation when applied directly on the point clouds, requiring multiple seconds, thus compromising real-time performance. 
In contrast, by directly producing the mesh using the simulated depth image from sensor hallucination, mesh reconstruction can be accelerated by more than two orders of magnitude; in our implementation, requiring less than $50\si{ms}$.
The main reason for this increase, apart from the reduction of the number of points considered, is due to the trivial computation of nearest neighboring points utilizing pixel-wise adjacency~\cite{izadi2011kinectfusion}.
In short, the point cloud is transformed into the world frame. 
For each set of four neighboring pixels defining a local image patch, up to two candidate triangles are generated by connecting the corresponding 3D points, and they are filtered based on inter-point distance thresholds, removal of duplicate faces, and angular consistency of the triangle normals.

For efficient convex decomposition, the Co-ACD~\cite{wei2022coacd} pipeline was employed which is computationally optimized and performs approximate convex decomposition maintaining a desired geometric accuracy.
To our experience, parameter tuning was necessary to produce results at a desired quality and computational efficiency, with respect to the simulated image resolution.
Nevertheless, finding parameters with reasonable performance was not particularly time consuming, and relative consistency was maintained across different scenes for the same simulated sensors.

Finally, the extracted convex polytopes were integrated in the Coal environment~\cite{coalweb}, which enables computationally efficient signed distance queries between convex volumes.

\subsection{Path Optimization}
The main motion planning of SHRUMS utilizes a path optimization approach, due to its established capacity to produce high-quality paths in real-time.
To provide continuous-time safety, the SHRUMS path-optimization process is inspired by AquaNav~\cite{xanthidis2020navigation}, a vision-based autonomous navigation pipeline, which has shown strong capacity for real-time 3D obstacle avoidance on agile underwater mobile platforms.
The objective function and constraints are enhanced with receding-horizon planning that minimizes the overall expected distance traveled to an arbitrarily distant global goal, similarly to the formulation from~\cite{amundsen2024rump}.

We form the optimization problem utilizing the objective function (\ref{eq:obj1}) and the constraints  (\ref{eq:obj2}).  For a pair of volumes $A_1$ and $A_2$, let $\operatorname{ch}(A_1, A_2)$ represent the convex hull of $A_1 \cup A_2$, and let $\operatorname{sd}(A_1, A_2)$ denote the signed distance between $A_1$ and $A_2$.  We write $H$ for the planning horizon considered, $d_{\rm safe}$ the minimum allowed safety distance between a robot state or transition to the nearby obstacles, $\lceil s \rceil$ for the 3D translational component of state $s$, and $s_n$ for the last state of path $S$ within the specified horizon.  
Then the problem can be defined as: 
    \begin{align}
        f(S) &= \displaystyle\min_{s_i \in S} \;
        \sum\limits_{i=1}^{n-1} \left \| s_{i+1}-s_{i} \right \|
        + w \left \| \lceil s_{g} \rceil - \lceil s_{n} \rceil \right \|,
        \label{eq:objective}  \\
        \textrm{s.t.}~\nonumber\\ 
        &\operatorname{sd}(\operatorname{ch}(A_{s_i}, A_{s_{i+1}}), A_{o}) \geq d_{\rm safe}, \,\,\, 1\leq i \leq n-1, o \in O
        \label{eq:col} \\
        & \left \| \lceil s_{n} \rceil - \lceil s_{1} \rceil \right \| = H. 
        \label{eq:gg}
    \end{align}

The first component of (\ref{eq:objective}) ensures minimal paths both in distance traveled and changes in orientation, while the second minimizes the expected distance between the local goal in the horizon and the global goal, with a weight $w$ regulating its contribution to the objective function.
The constraint (\ref{eq:gg}) allows the local goal to move dynamically during optimization at a distance equal to the horizon $H$.
Constraint~(\ref{eq:col}) guarantees continuous time safety, avoiding collisions during the entire trajectory with a swept-out volume formulation, by computing the convex hull between each pair of consecutive states, and then enforcing a minimum positive signed distance between the swept-out volume and the surrounding obstacles.
The optimization problem is solved with a sequential least-squares quadratic programming (SLSQP) solver~\cite{Kraft1988SLSQP}.
An example of a solution is shown in Figure~\ref{fig:beauty}d.

For this kind of optimization, warm-starting initializations are often used, taking one of two forms: 1) linear interpolation towards the global goal for a length $H$, or 2) utilizing the previous computed path. 
The first option may avoid potential local minima in the absence of the global planner, but may require more iterations to find a solution. 
The second option initializes the path using the previous solutions, thus it can provide valid solutions faster.
We utilized linear interpolation for the initial solutions and then previous solution to warm-start new ones, though other policies could be developed.

\subsection{Path Following}
A simple path follower was designed to provide references to the controllers enabling the robot to follow the computed path.
The computed reference is at a desired look-ahead distance from the current position, while the reference traverses along the computed path, adjusting desired 3D orientations with interpolation towards the next state to be achieved.

\section{Experimental Validation}
\label{sec:exp}
 \begin{figure}[ht]
\centering
\leavevmode
\newcommand{\w}{0.22\textwidth}
\begin{tabular}{cc}
 \subfigure[]{\includegraphics[width=\w]{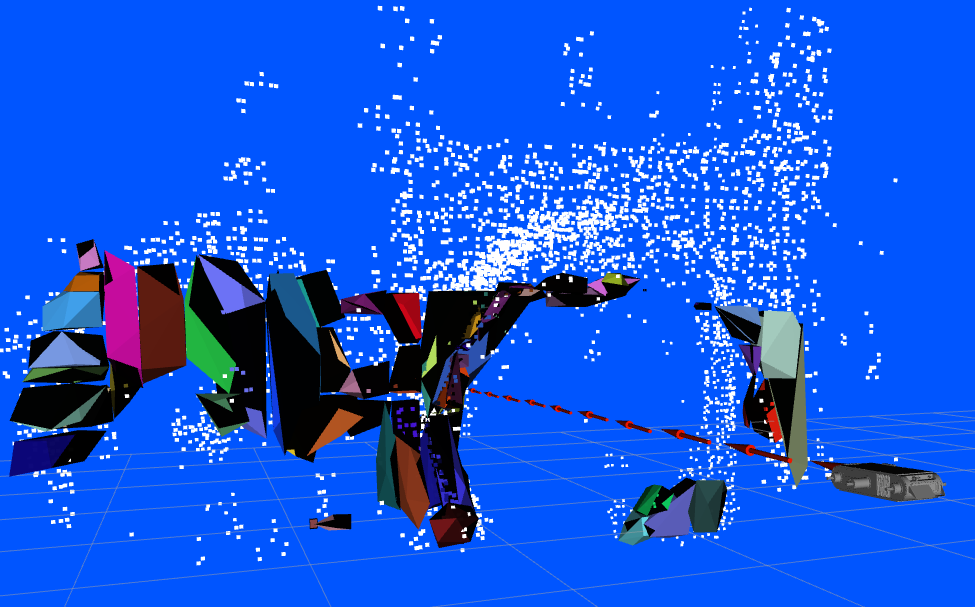}\label{fig:poolMap1}}&
\subfigure[]{\includegraphics[width=\w]{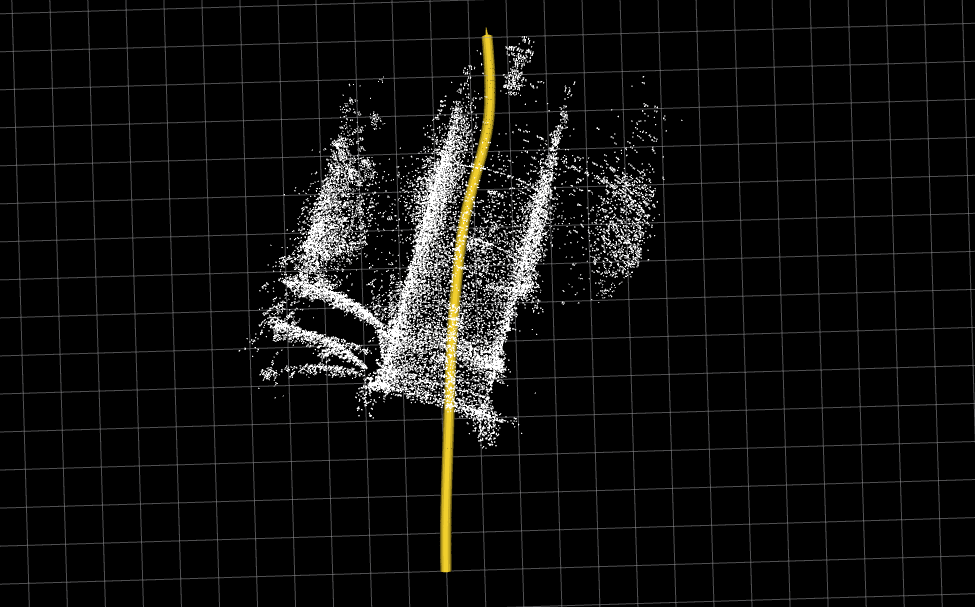}\label{fig:poolMap4}}\\
 \subfigure[]{\includegraphics[width=\w]{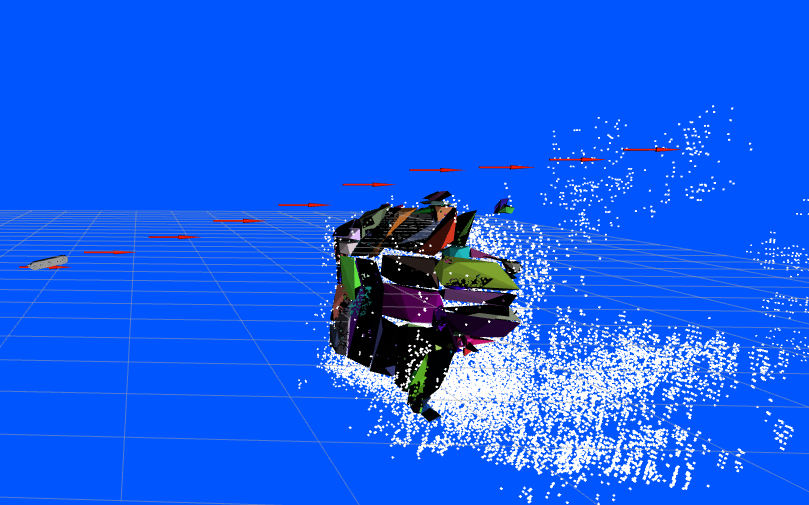}\label{fig:poolMap1_}}&
\subfigure[]{\includegraphics[width=\w]{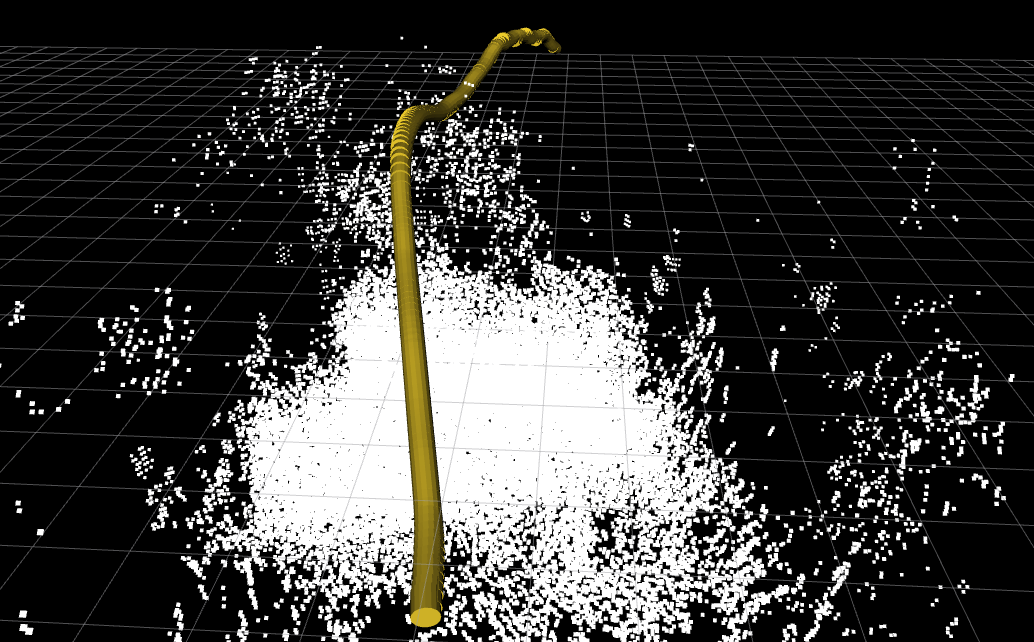}\label{fig:poolMap4_}}\\
 \subfigure[]{\includegraphics[width=\w]{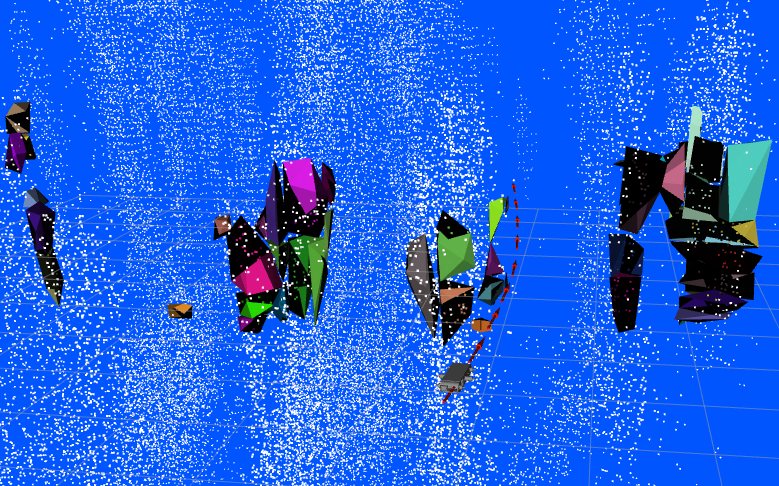}\label{fig:pool1i}}&
\subfigure[]{\includegraphics[width=\w]{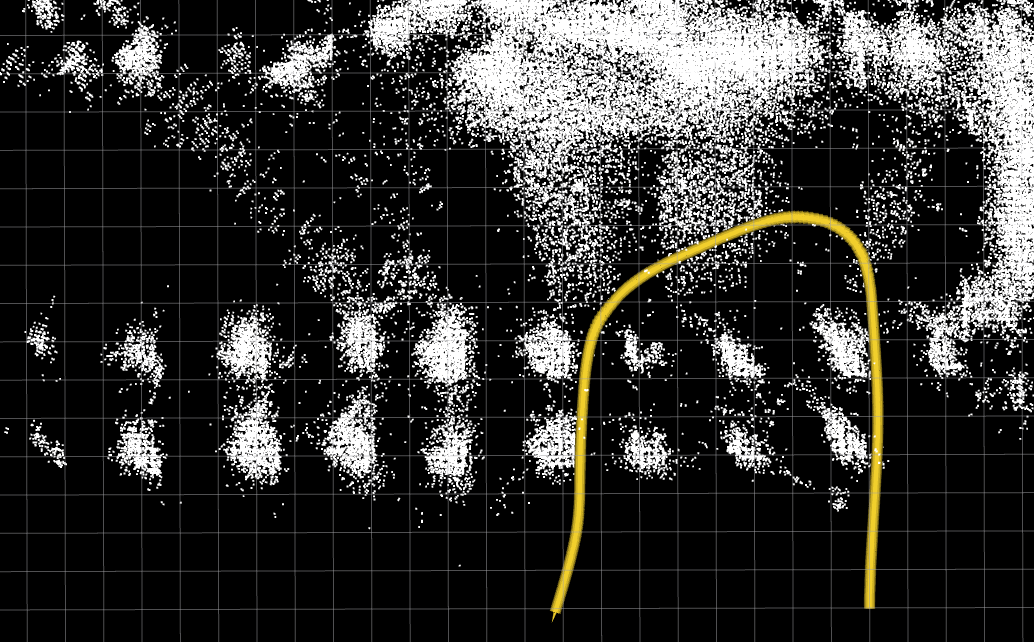}\label{fig:pool4i}}
\end{tabular}
   \caption{Qualitative results for the Container, Car, and Pillars environment (rows)). Planning instance for each at the C1 configuration showing the convex polytopes representation and the path with red arrows in (a,c,e), while the yellow the trajectory of the robot for the C3 configuration in (b,d,f).}
   \label{fig:test_cases}
\end{figure}
The validation of SHRUMS in this study includes two main elements: (i) tests real data from Water Linked's Sonar 3D-15 coupled with simulated dynamics and control emulating the Aqua2 hexapod UAV~\cite{dudek2007aqua} for the control and realistic dynamics, and (ii) real deployments of a BlueROV in both pool and field conditions.

%

To perform our testing, we exploited sensor hallucination with its unique decoupling between the data gathered from the real sensors of the robot and the simulated sensor, which informed autonomous navigation.
Thus, map segments were used as created simultaneously by the state-estimation, and the robot was tasked with navigating in these maps using the simulated sensor, informed directly by the point cloud of the local map.
We tested three different sensor configurations, mounted in the same position as the original sensor:
\begin{itemize}
    \item C1: Downsampling the resolution of the original sensor while maintaining its field of view and decreasing its range to 10\si{m}, resulting in an image of $100\times 25$ pixels and FOV of $100^\circ \times 25^\circ$.
    \item C2: Matching the resolution of the original sensor, its field of view, and range to 15\si{m,} resulting in an image of $200\times 50$ pixels and FOV of $100^\circ \times 25^\circ$.
    \item C3: Defining a different sensor providing better vertical view than the original sensor, resulting in an image of $100\times 50$ pixels and FOV of $100^\circ \times 50^\circ$, though significantly decreasing the range to only 5\si{m}.
\end{itemize}

\subsection{Simulations Driven by Real 3D Sonar Data}
Three datasets from three different locations were used to produce maps in real-time with KISS-ICP~\cite{vizzo2023ral}, including an underwater cylindrical container (Container), a submerged car in a benthic environment (Car), and a complex dock structure with pillars (Pillars).
In all datasets, the visibility conditions were poor, challenging vision-based pipelines (Figures~\ref{fig:beauty}a and ~\ref{fig:median}b).  
In contrast, consistent local maps were produced with the 3D sonar, capturing the challenging geometry of the structures to a significant geometric detail.
The maps were constructed in real-time, matching the frequency of the sensor at $5$-$6$\si{Hz}, and in our experience without negatively impacting the performance of the SHRUMS pipeline.
Our implementation was run on a laptop with an Intel Core i9-13900 at 5.6 GHz and 16 GB RAM.

The horizon was set to 10\si{m}, the path resolution to 1 state per meter, the look-ahead distance to $1.2$, and the robot's velocity to $0.3$ m/s. 
The parameters for convex decomposition in CoACD were set as follows: threshold to $0.034$, mcts\_max\_depth  to $1$, preprocess\_resolution  to $25$, mcts\_nodes  to $10$, and mcts\_iterations  to $60$.

Example instances from our experiments are shown in Figure~\ref{fig:test_cases}, with the Container and Car being tested with a pass through objective with a single goal, while the Pillars including 3 goals to enable multigoal navigation inside the confined structure after entering and before exiting.
No collisions occurred with any of the environments as represented with point clouds, though there were few instances that the robot collided with few points of the map.
In all such cases, these points were outliers. 
This indicates further robustness and stems directly as an attribute of the sensor hallucination applied in our implementation, with such isolated points filtered by the simulated depth image.
Regarding the configurations, C1 and C2 performed similarly in the testing environments, with C2, in general, providing slightly shorter paths, given the larger horizon and image resolution.
C3 in all cases provided slightly longer paths and showed a more reactive performance due to the shorter range, especially vertically, due to the increased vertical FOV.

\begin{table}[t]
\begin{tabular}{c|ccccc}
             & \multicolumn{5}{c}{\textbf{SHRUMS}}                                                                                                                                                                                                                                                                                                                                                                                             \\ \hline
\textbf{Map} & \multicolumn{1}{c|}{\textbf{\begin{tabular}[c]{@{}c@{}}Sensor\\ Config.\end{tabular}}} & \multicolumn{1}{c|}{\textbf{\begin{tabular}[c]{@{}c@{}}Sensor\\ Simulation\end{tabular}}} & \multicolumn{1}{c|}{\textbf{\begin{tabular}[c]{@{}c@{}}Mesh\\ Recon.\end{tabular}}} & \multicolumn{1}{c|}{\textbf{\begin{tabular}[c]{@{}c@{}}Convex\\ Decomp.\end{tabular}}} & \textbf{\begin{tabular}[c]{@{}c@{}}Path\\ Opt.\end{tabular}} \\ \hline
             & \multicolumn{1}{c|}{C1}                                                                & \multicolumn{1}{c|}{0.0022}                                                                  & \multicolumn{1}{c|}{0.0025}                                                            & \multicolumn{1}{c|}{0.7701}                                                               & 0.6808                                                           \\ \cline{2-6} 
Container    
             & \multicolumn{1}{c|}{C2}                                                                & \multicolumn{1}{c|}{0.0029}                                                                  & \multicolumn{1}{c|}{0.0446}                                                            & \multicolumn{1}{c|}{0.8008}                                                               & 0.7268                                                          \\ 
             \cline{2-6} & \multicolumn{1}{c|}{C3}
             & \multicolumn{1}{c|}{0.0035}
             & \multicolumn{1}{c|}{0.0299}
             & \multicolumn{1}{c|}{0.7503} & 0.4035 \\
             \hline
             & \multicolumn{1}{c|}{C1}                                                                & \multicolumn{1}{c|}{0.0062}                                                                  & \multicolumn{1}{c|}{0.0143}                                                            & \multicolumn{1}{c|}{0.9606}                                                               & 0.1930                                                          \\ \cline{2-6} 
Car      & \multicolumn{1}{c|}{C2}                                                                & \multicolumn{1}{c|}{0.0051}                                                                  & \multicolumn{1}{c|}{0.0337}                                                            & \multicolumn{1}{c|}{1.0471}                                                               & 0.2079                                                          \\ \cline{2-6} 
             & \multicolumn{1}{c|}{C3}                                                                & \multicolumn{1}{c|}{0.0054}                                                                  & \multicolumn{1}{c|}{0.0327}                                                            & \multicolumn{1}{c|}{0.9589}                                                               & 0.4215                                                          \\ \hline
             & \multicolumn{1}{c|}{C1}                                                                & \multicolumn{1}{c|}{0.0113}                                                                  & \multicolumn{1}{c|}{0.0160}                                                            & \multicolumn{1}{c|}{1.3780}                                                               & 0.0717                                                          \\ \cline{2-6} 
Pillars          & \multicolumn{1}{c|}{C2}                                                                & \multicolumn{1}{c|}{0.0121}                                                                  & \multicolumn{1}{c|}{0.0334}                                                            & \multicolumn{1}{c|}{1.6312}                                                               & 0.0700                                                          \\ \cline{2-6} 
             & \multicolumn{1}{c|}{C3}                                                                & \multicolumn{1}{c|}{0.0095}                                                                  & \multicolumn{1}{c|}{0.0199}                                                            & \multicolumn{1}{c|}{1.1514}                                                               & \multicolumn{1}{c}{0.0766}                                    
\end{tabular}
\caption{Average times of different stages of SHRUMS (s)}
\label{tab:times}
\end{table}

Table~\ref{tab:times} shows the computational time for the different stages of SHRUMS, providing strong indications for real-time performance in the underwater domain. 
Sensor simulation required on average at most 10\si{ms} and mesh reconstruction 40\si{ms}.
The most computationally expensive operation was convex decomposition, on average reaching at most around 1.6\si{s}, although parameter tuning 
could further accelerate the map updates.
Then, the motion planning, which provides resolution-independent continuous-time safety and considers the undiscretized geometry of the environment, remained less than $1$\si{s}, while shorter horizons could improve efficiency further.
Finally, it is worth noting that the mesh reconstruction process on the simulated depth image was more than two orders of magnitude faster than the standard Open3D's implementation of the Poisson mesh reconstruction performed on the unorganized point clouds of the local maps, requiring on average $8.5182$\si{s} for the Container, $8.4734$\si{s} for the Car, and $7.4097$\si{s} for the Pillars. 

\subsection{Real Deployments}

 \begin{figure}[ht]
\centering
\leavevmode
\newcommand{\w}{0.22\textwidth}
\begin{tabular}{cc}
 \subfigure[]{\includegraphics[width=\w, trim=7cm 2cm 7cm 5cm, clip=true]{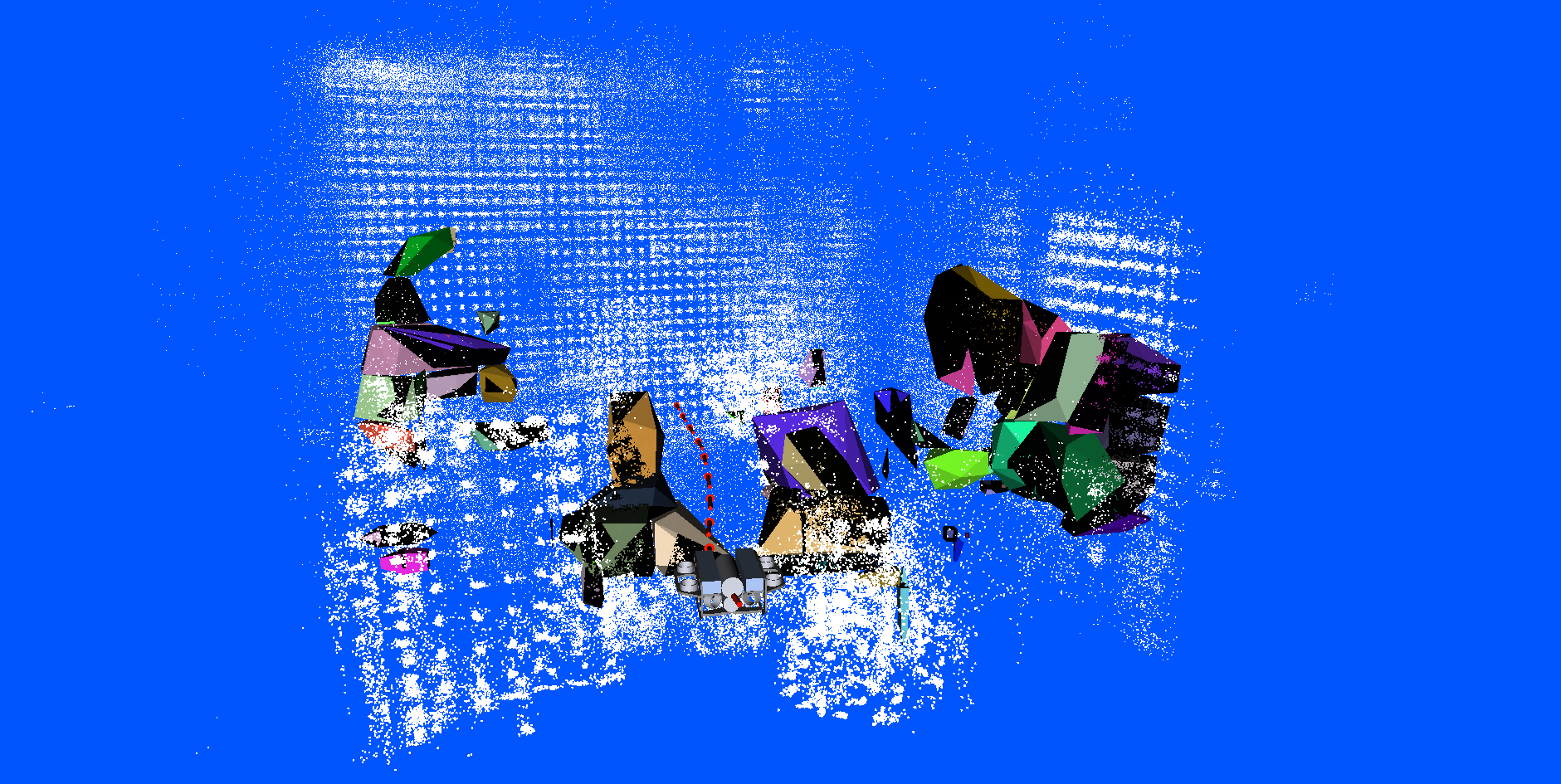}\label{fig:pool1}}&
 \subfigure[]{\includegraphics[width=\w]{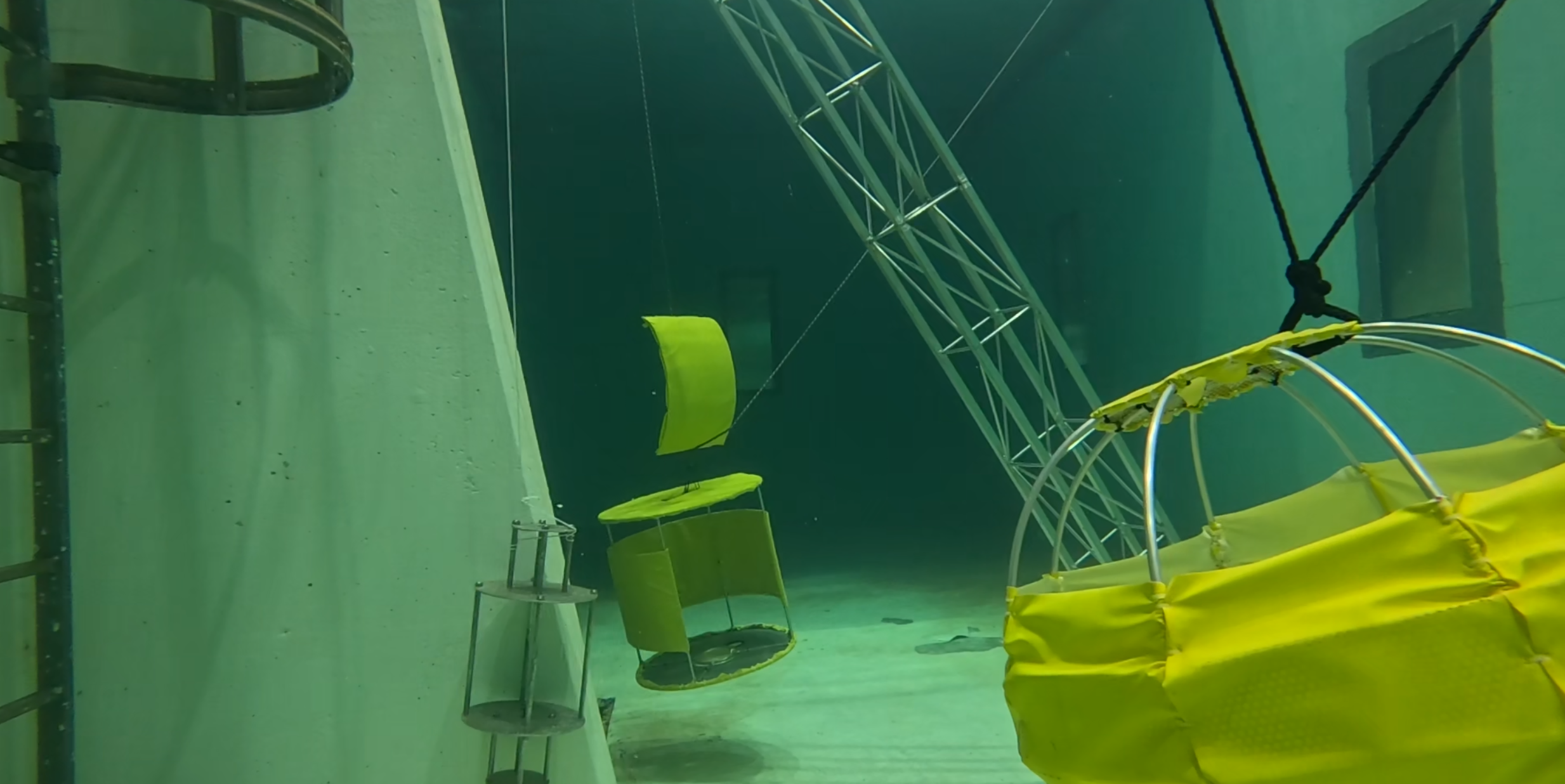}\label{fig:pool3}}\\
 \subfigure[]{\includegraphics[width=\w, trim=10cm 2cm 4cm 5cm, clip=true]{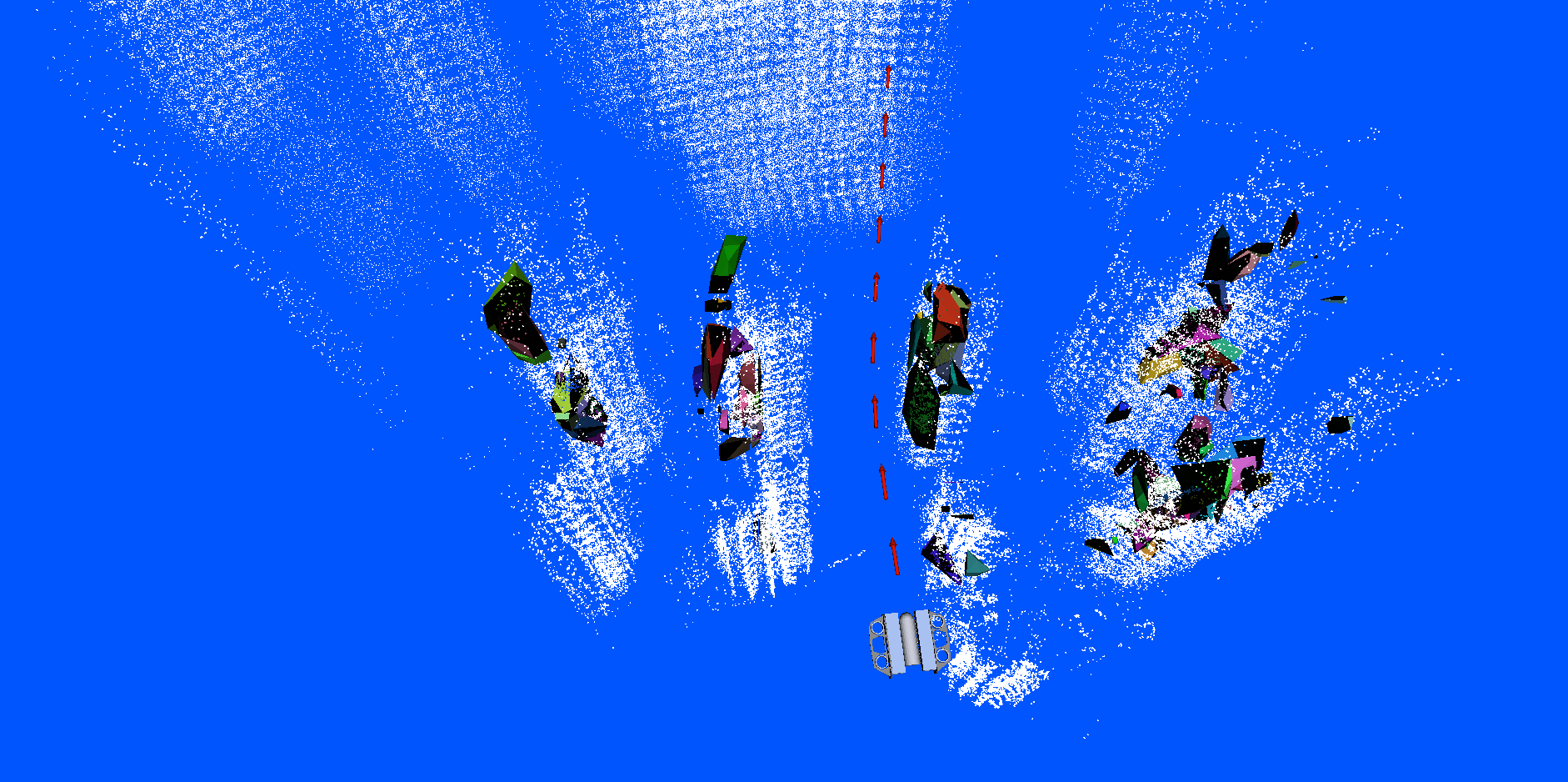}\label{fig:pillar1}}&
\subfigure[]{\includegraphics[width=\w]{figs/pillars_cam.png}\label{fig:pillar3}}\\
 \subfigure[]{\includegraphics[width=\w]{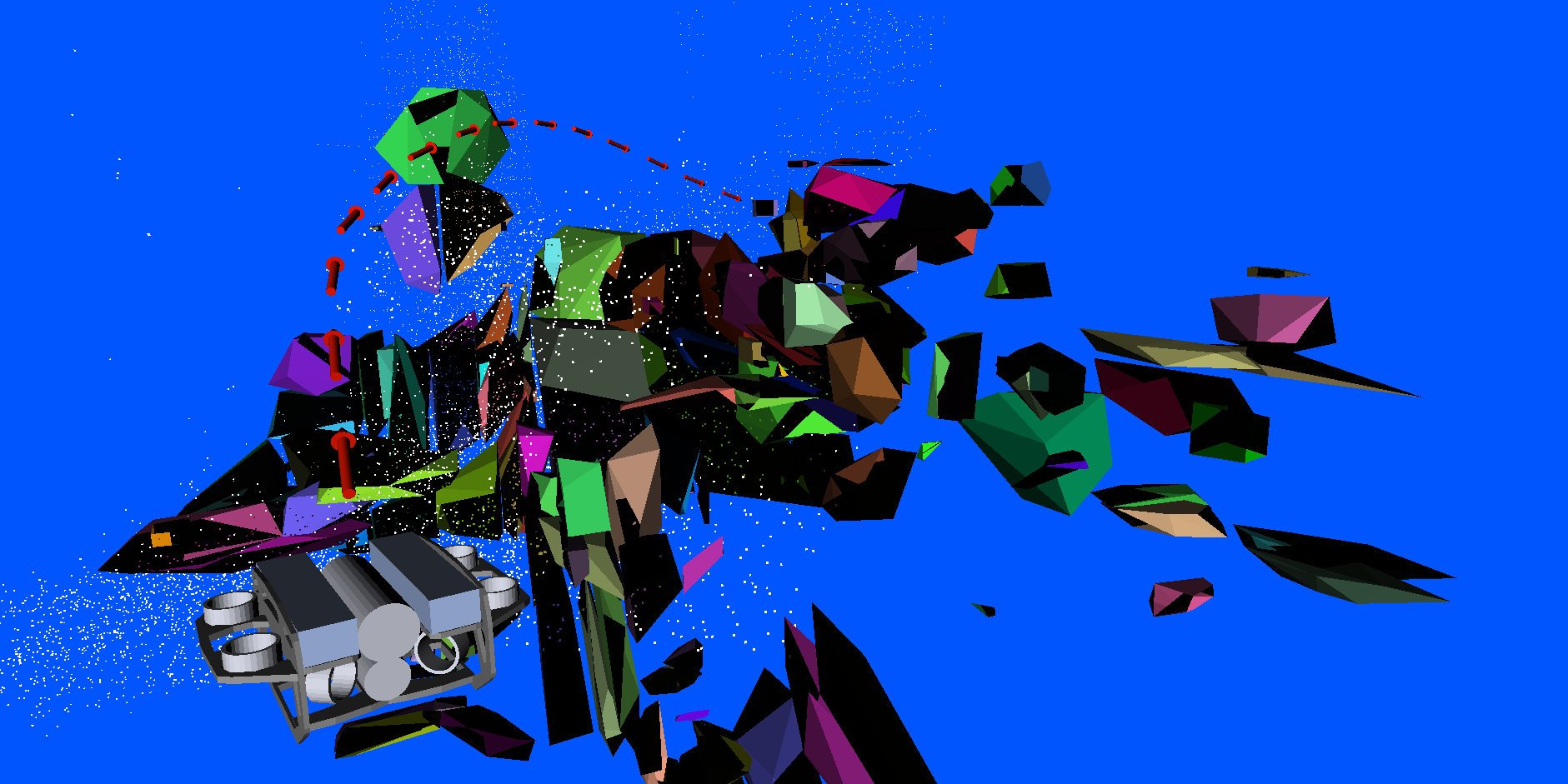}\label{fig:elli1}}&
\subfigure[]{\includegraphics[width=\w]{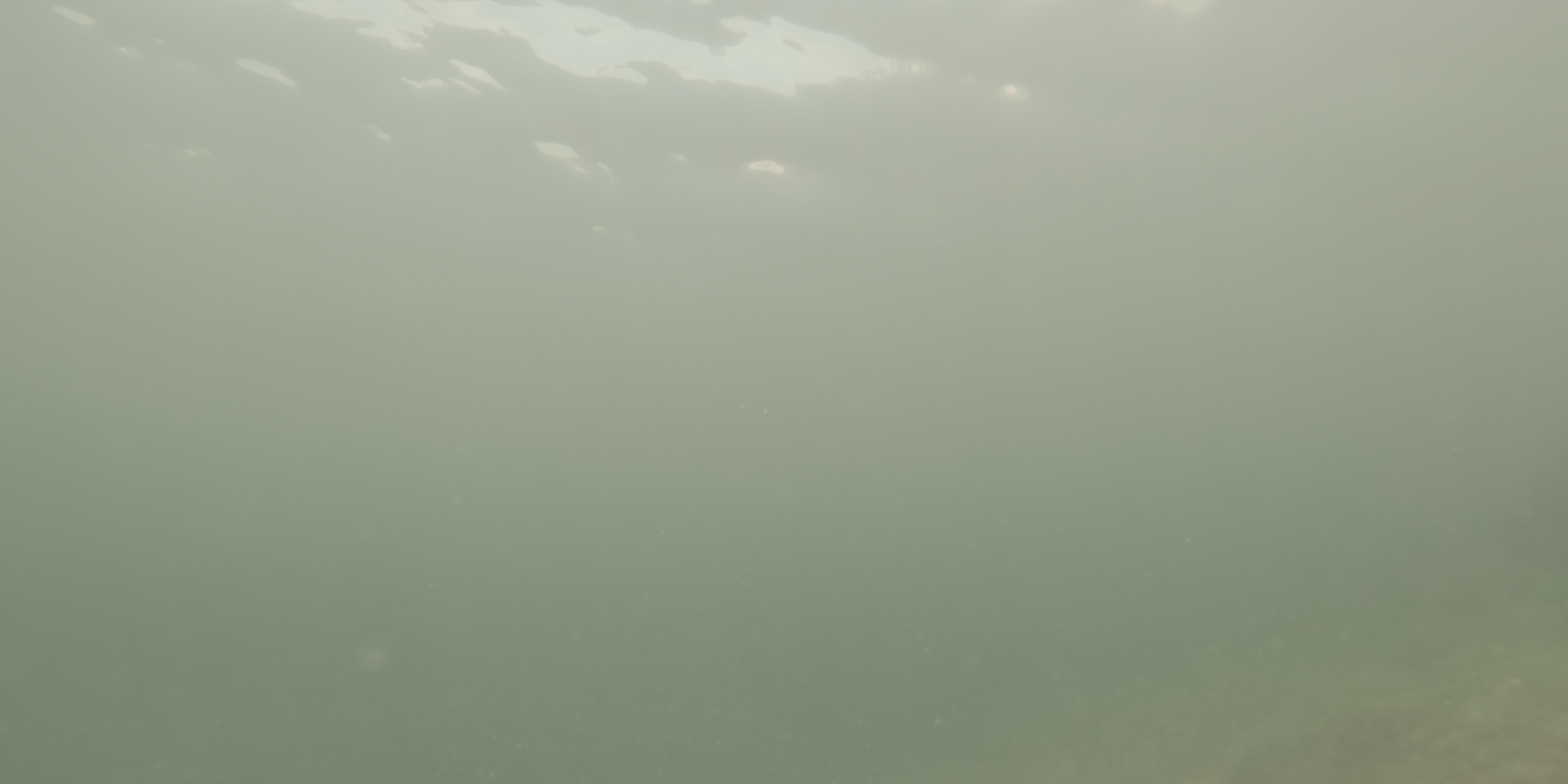}\label{fig:elli3}}
\end{tabular}
   \caption{Qualitative results from  E1, E2, and E3 deployments (rows). Planning instances during operation shown in (a,c,e), while  a representative visual inputs in (b,d,f).}
   \label{fig:field_exp}
\end{figure}
SHRUMS was also tested on a tethered BlueROV heavy configuration employed with a forward looking Water Linked Sonar 3D-15, and a bottom-facing Water Linked DVL A-50.
To test the robustness of SHRUMS in real-conditions, the sonar scans were registered directly relying an error-state Kalman filter utilizing the on-board IMU, magnetometer, and DVL readings at 100\si{Hz}, with a 10\si{s} sliding window.
The robot utilized line-of-sight guidance with an agile learning-based controller~\cite{fosso2025sim2swim}.
All computations, including state-estimation, mapping, policy inference for controls, and the proposed pipeline, were performed remotely on a single laptop with an Intel core Ultra9 275HX processor at 5.4 GHz, without exploiting GPU parallelism. 
The path resolution was set to $0.5$\si{m} to enforce more adaptive performance to counter moving objects in the surroundings, potential registration drifts, and, especially, currents and disturbances due to on-going marine traffic and significant tidal conditions. 

Three test-cases were constructed testing the capacity of SHRUMS in three different challenges: E1) A complex cluttered pool environment involving walls and semitransparent aluminum obstacles, intended as a proxy for industrial settings or disaster response scenarios, E2) The Pillars environment, in the same location as the simulation dataset,  with added wavezone disturbances and currents, intended to uncover the sim-to-real gap, and finally, E3) a cluttered port environment, aimed to test the capacity for 3D navigation while encountering marine life and currents.
Multiple successful runs were performed in each scenario. Figure~\ref{fig:field_exp} summarizes these deployments.
Note that in environments E2 and E3, visibility was extremely poor, exceeding the typical range of applicability of vision-focused SLAM pipelines.
%

\section{Conclusion}
\label{sec:conc}
A new pipeline was proposed for robust autonomous underwater 3D navigation, called SHRUMS. 
SHRUMS introduces, for the first time, a comprehensive process for incorporating compact 3D sonar data to perceive and enable navigation in complex environments with arbitrary visibility conditions.
Additionally, it introduced the concept of sensor hallucination for reducing computational expense, preserving the map geometry without volumetric discretization, while also improving sensing quality from the on-board sensors in a decoupled way.
Future work will focus on implementation improvements on the presented pipeline with further investigation on the capacity of compact 3D sonars for autonomous navigation and inspection. 
We are also particularly interested in exploring the opportunities of sensor hallucination towards improving inspection~\cite{xanthidis2024resivis} or navigation techniques, transfer learning, and out-of-body sensing. 

\section{Acknowledgments}
\label{sec:ackn}
The authors would like to thank Dr. Nikhil Khedekar and Dr. Mihir Dharmadhikari (NTNU) for early discussions, Dr. Alberto Quattrini Li (Dartmouth) for assistance in sensor integration, and  Water Linked for their support, including access to their hardware and datasets during testing.
\bibliographystyle{IEEEtran}
\bibliography{./IEEEabrv,refs}


\end{document}